\documentclass[letterpaper, 10 pt, conference]{ieeeconf}  
\IEEEoverridecommandlockouts                              
\usepackage{packages}

\title{\LARGE\bf 

CLAP: Clustering to Localize Across $n$ Possibilities, \\
A Simple, Robust Geometric Approach in the Presence of Symmetries 

}
\author{Gabriel I.~Fernandez$^{1, 2, \dagger}$, Ruochen Hou$^{1, \dagger}$, Alex Xu$^{1}$, Colin Togashi$^{1, 2}$, and Dennis W.~Hong$^{1}$
\thanks{$^{\dagger}$ Denotes equal contribution.}
\thanks{
$^{1}$Robotics and Mechanisms Laboratory (RoMeLa), Department of Mechanical and Aerospace Engineering, University of California, Los Angeles, CA 90095, USA.
        {\tt\small \{gabriel808, houruochen, ax90l0tl, ctogashi, dennishong\}@ucla.edu}}
\thanks{
$^{2}$VEQdrive, robotics \& machine learning startup focused on automation.
}
}

\begin{document}
\maketitle
\thispagestyle{empty}
\pagestyle{empty}

\begin{abstract}

In this paper, we present our localization method called CLAP, Clustering to Localize Across $n$ Possibilities, which helped us win the RoboCup 2024 adult-sized autonomous humanoid soccer competition. Competition rules limited our sensor suite to stereo vision and an inertial sensor, similar to humans. In addition, our robot had to deal with varying lighting conditions, dynamic feature occlusions, noise from high-impact stepping, and mistaken features from bystanders and neighboring fields. Therefore, we needed an accurate, and most importantly robust localization algorithm that would be the foundation for our path-planning and game-strategy algorithms. CLAP achieves these requirements by clustering estimated states of our robot from pairs of field features to localize its global position and orientation. Correct state estimates naturally cluster together, while incorrect estimates spread apart, making CLAP resilient to noise and incorrect inputs. CLAP is paired with a particle filter and an extended Kalman filter to improve consistency and smoothness. Tests of CLAP with other landmark-based localization methods showed similar accuracy. However, tests with increased false positive feature detection showed that CLAP outperformed other methods in terms of robustness with very little divergence and velocity jumps. Our localization performed well in competition, allowing our robot to shoot faraway goals and narrowly defend our goal.
\end{abstract}


\section{Introduction}
\label{sec:introduction}
\begin{figure}[t!]
 \begin{subfigure}[]{0.95\linewidth}
     \centering
     \includegraphics[width=\textwidth]{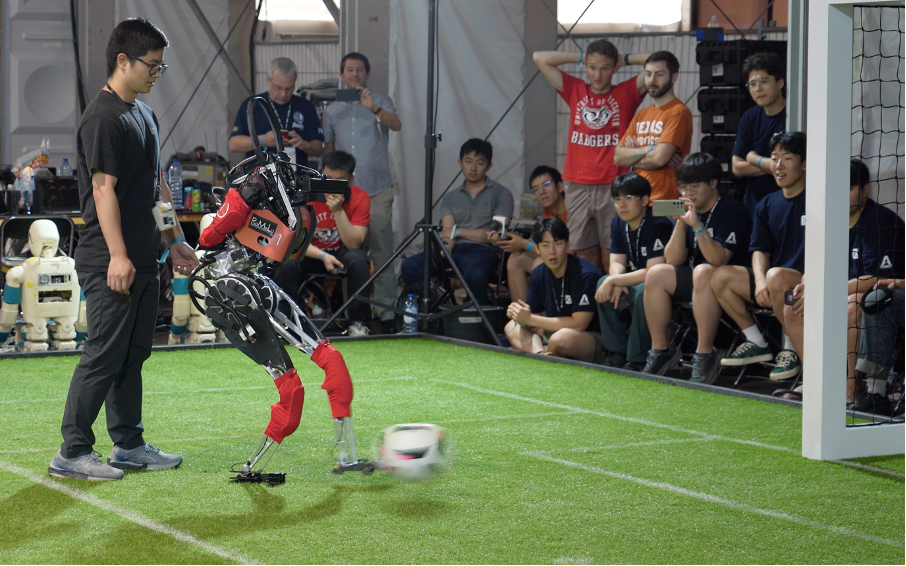}
     \caption{}
     \label{fig:main}
 \end{subfigure}
 \begin{subfigure}[]{0.95\linewidth}
     \centering
     \includegraphics[width=\textwidth]{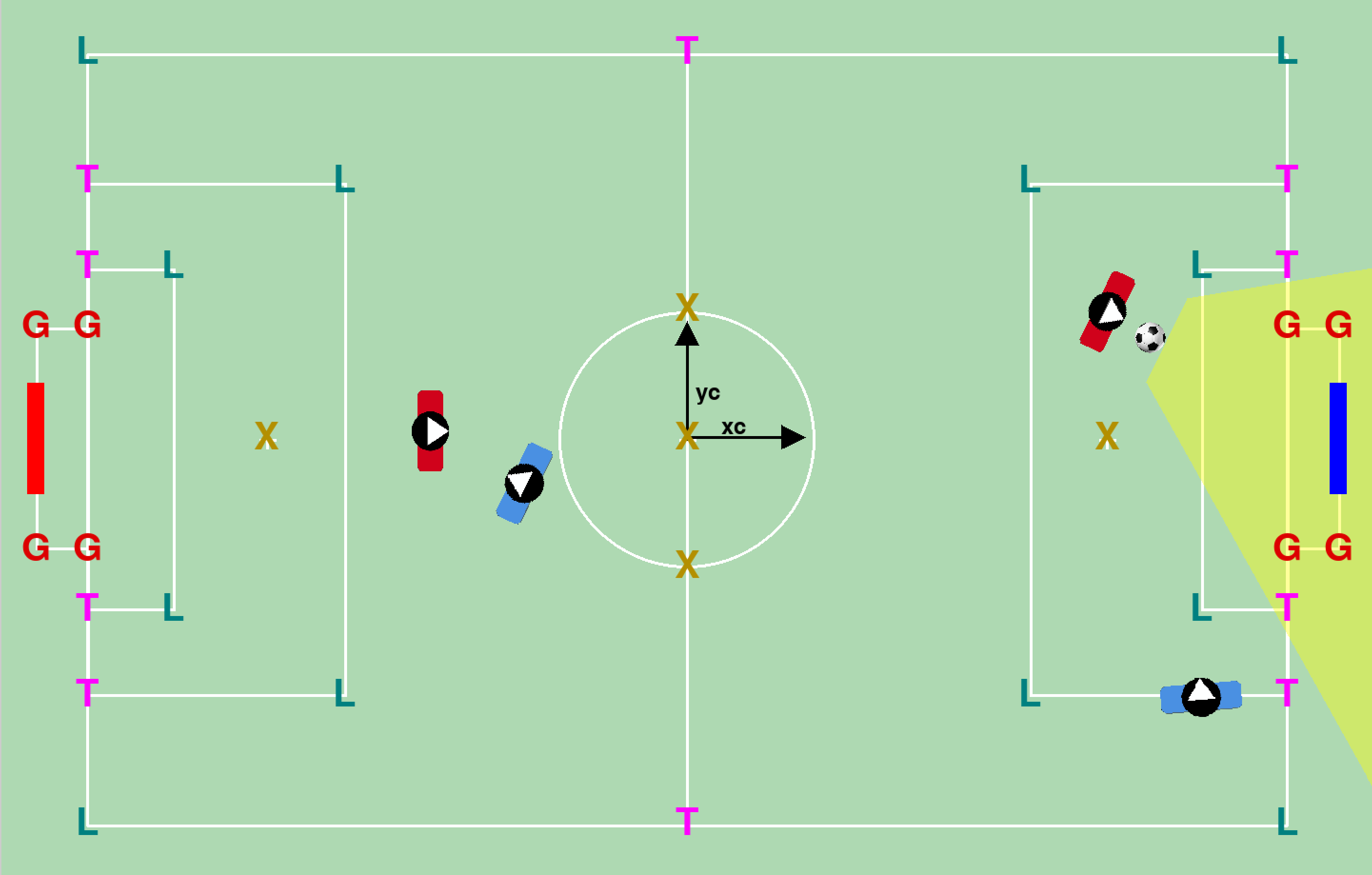}
     \caption{}
     \label{fig:landmarks}
 \end{subfigure}
\caption{(a) An image from the Robocup championship match with our robot, Artemis, shown with red bands taking a shot on goal. For safety measures a person dressed in black with an emergency stop follows our robot.
(b) The state of game just for illustrative purposes assuming no noise. Field features are defined by: corners denoted with a \textit{green L}, T-intersections with a \textit{pink T}, goal posts with a \textit{red G}, and crosses with a \textit{yellow X}. The global referential frame is drawn in \textit{black} lines in the center of the field whose axes are labelled \textit{yc}, \textit{xc}. The length and width of the field of play is 14 m by 9 m. The field also consists of a meter of extra space outside the field lines for throw-ins, corner kicks, etc. The players of each team, red and blue, are depicted with team color rectangles with a black circle and triangle to indicate their orientation. The \textit{yellow} highlight indicates the area captured by the camera.}
\end{figure}

Every year, the Robocup Federation hosts a humanoid soccer competition in hopes of one day playing a live match of robots versus humans. To ensure a fair match, rules are put in place such that robots must be able to play autonomously, be of similar physiological proportions to a human, and only be equipped with sensors that have biological equivalents. For example, cameras are allowed because they are analogous to eyes, but magnetometers cannot be used. In addition, robots must play on a standardized green turf field marked with white lines for field boundaries, penalty areas, and goals, as shown in \cref{fig:landmarks}.

\begin{figure*}[t!]
    \centering
    \includegraphics[width=0.9\linewidth]{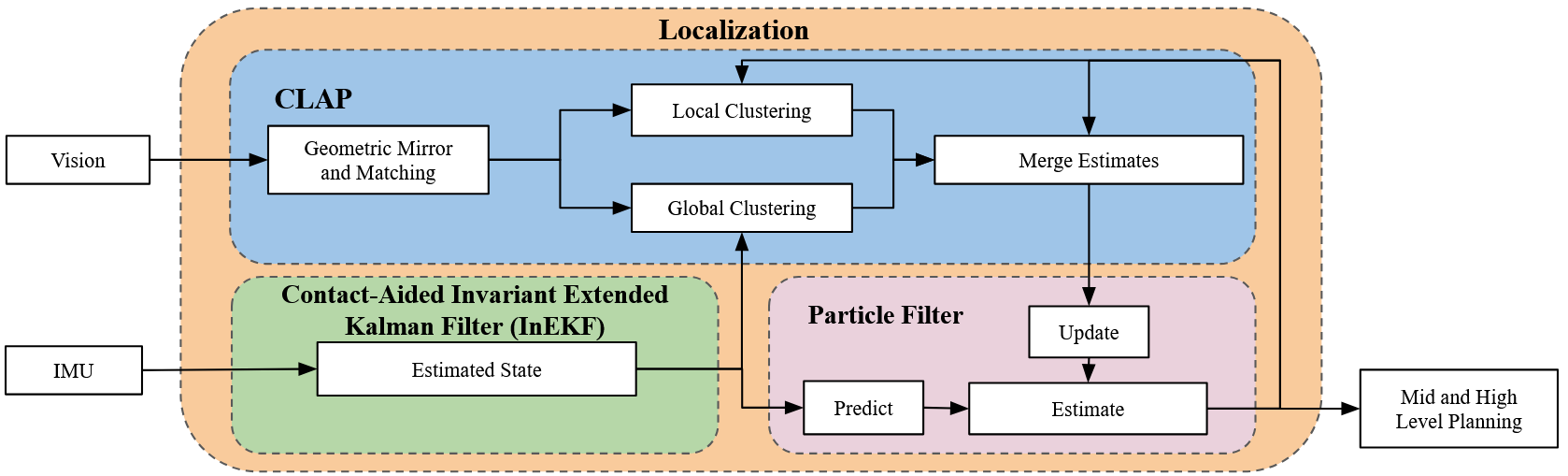}
    \caption{Flowchart of the full localization framework in \textit{orange}, CLAP in \textit{blue}, IMU state estimator in \textit{green}, and particle filter in \textit{pink}. The IMU state estimator uses a Contact-Aided Invariant Extended Kalman Filter (InEKF)}
    \label{fig:framework}
\end{figure*}

For any effective soccer strategy, it is essential that the robot knows its localized state on the field of play at all times. Both position $(x, y)$ and angle $\theta$ relative to a defined coordinate system must be known for high-level planning, for example, determining whether to shoot or not. 


Furthermore, the localization algorithm needs to be very robust to external noise and limited information. A typical match is a very dynamic environment with multiple robots, human handlers, and referees all moving on the field at the same time. In addition, lighting conditions vary greatly throughout the day with sudden flashes coming from cameras in the crowd. Landmarks can be occluded or even misclassified depending on the current view. Games in nearby fields also run concurrently and can cause further misdetections. The robot even experiences large vibrations when it takes a step, not to mention when it runs into another robot, blurring images and adding to noise to the IMU.

Localization must also solve quickly enough in order to prevent delays from propagating through the system. An accurate, yet highly delayed estimate would cause any trajectory tracking to overshoot drastically. This would result in collisions with other robots, the goal posts, or even walking off into the crowd. As was discovered, the rate at which a path planning controller could be run at was directly capped by the delay of the localization algorithm \cite{hou2025path}. Thus, the localization rate was a factor that could not be ignored.

The proposed solution, CLAP, uses a simple geometric approach to estimate states that naturally form clusters when relative camera observations are consistent with the a priori known global map of the field features. In other words, the algorithm selects the position and orientation on the field that most closely matches landmarks with their predicted positions. Since geometric calculations are relatively simple to compute and parallelize, CLAP offers an efficient and accurate solution to the localization problem. Meanwhile, clustering in both the robot and field frames allows the algorithm to ignore clear outliers and be robust to misdetections. Since it only uses vision-based landmarks, state estimates from other sources are fused to give an estimate between camera frame updates. 

The following is a summary of contributions in this paper:
\begin{enumerate}
    \item Introduced a novel localization method using clustering which is fast and robust to large disturbances, 
    \item Analyze nuisances of such a method, and 
    \item Compare its accuracy and robustness performance with alternative localization methods and demonstrate its capabilities in an international soccer competition.
\end{enumerate}

The rest of the paper is organized into the following: \cref{sec:algorithm} explores the CLAP algorithm in depth, giving intuition. \cref{sec:results} discusses simulation results, and comparisons with other localization methods. Finally, \cref{sec:conclusion} offers closing thoughts.


\section{CLAP Localization} 
\subsection{Related Work}
Several solutions for localization on a soccer field were considered; however, they failed to perform well given the aforementioned criteria. Visual odometry-based methods \cite{oriolo2012vision}, which rely on tracking visual features between consecutive frames, were too sensitive to moving objects in the environment. Learning-based methods were also tested \cite{kendall2015posenet}, but relied too heavily on collecting detailed data in order to perform well. Meanwhile, Monte-Carlo-based Localization (MCL) \cite{kim2023enhancing,thrun2005probabilistic,hong2009robust,hornung2010humanoid,almeida2017vision,nagi2014vision} was computationally heavy and slow, specifically when the number of particles was increased for accuracy. Among the Monte Carlo methods, augmented Monte Carlo Localization (aMCL)—a variant of MCL used by the Hanyang University Robocup team—has shown greater accuracy than visual SLAM methods \cite{kim2023enhancing}.

We also simultaneously developed and tested an optimization-based algorithm called Iterative Landmark Matching (ILM). ILM is based on the Iterative Closest Point (ICP) algorithm, but has been shown to be more robust than ICP and faster than aMCL, while achieving even higher accuracy \cite{ICP,hou2025localization}. Both ILM and aMCL will be compared with CLAP in \cref{sec:results}.

\subsection{Overview}
\label{sec:algorithm}
CLAP is comprised of several key components, see \cref{fig:framework}. The first and most unique is the geometric matching using vision-based landmarks. It takes a priori known field features to estimate possible states and then uses unsupervised clustering to determine an estimate closest to the true state. The particle filter and Contact-Aided Invariant Extended Kalman Filter (InEKF) will be discussed further in \cref{sec:other_components}.


\begin{figure}[t!]
    \centering
    \includegraphics[width=0.9\linewidth]{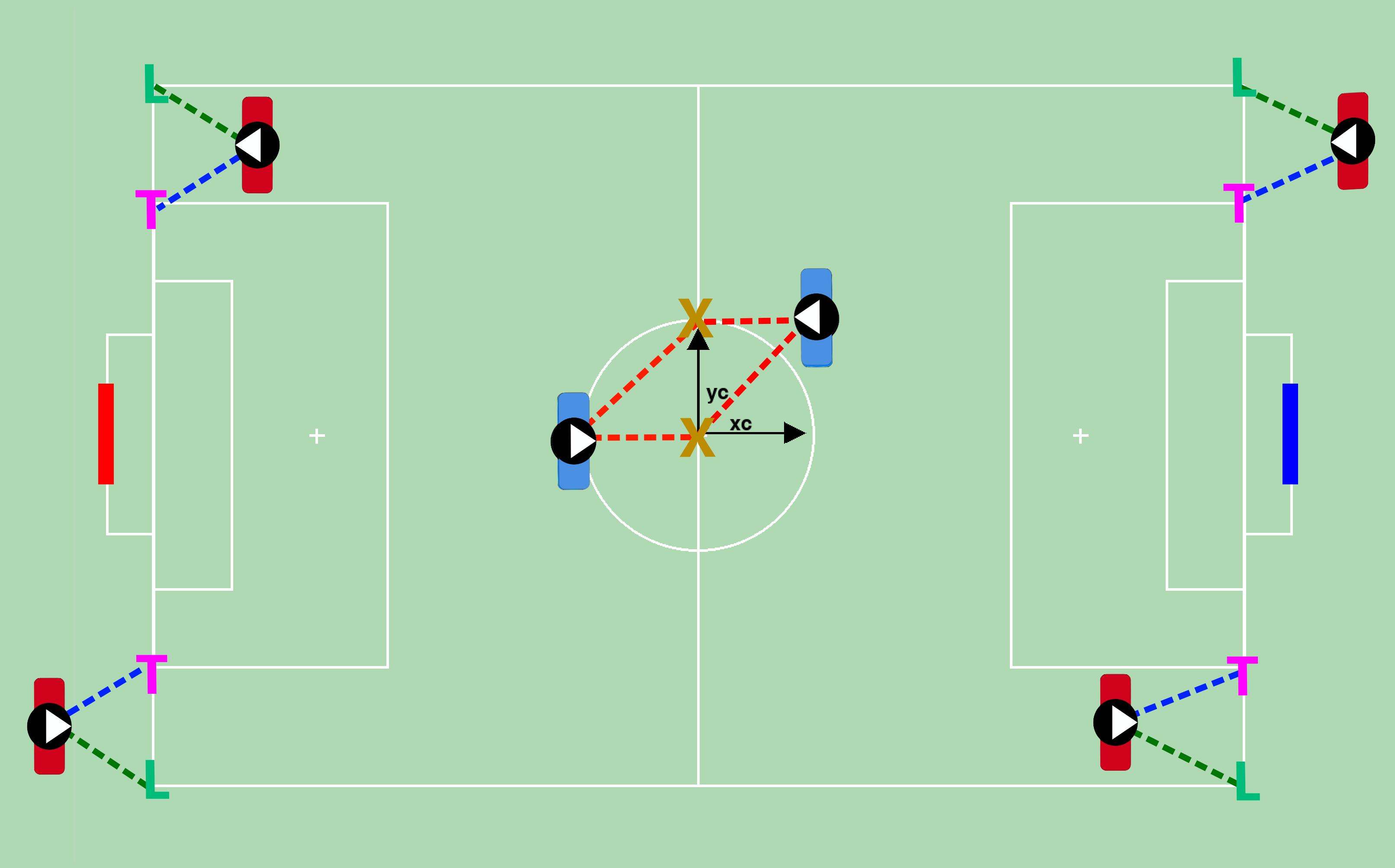}
    \caption{In this figure the \textit{red} player are matched with \textit{T} and \textit{L} pairs that share the same relative distance between markers. Note that in order to keep the spatial relationship in the local frame that it cannot be simply mirrored and only one solution exists. The \textit{green} dotted line always touches the \textit{L} and the \textit{blue} dotted line the \textit{T}. Also notice that the \textit{red} robots are far from one another. Conversely, the \textit{blue} robots in the center can be on either side of the \textit{X} since the landmarks are indistinguishable.}
    \label{fig:mirroring}
\end{figure}

\subsection{Geometric Mirroring}
\label{sec:geo_mirror}
Since most soccer behaviors only require the robot's pose $(x_{robot},y_{robot}, \theta_{robot})$ relative to a frame on the field, the problem can be simplified to 2D by projecting the landmarks seen by the camera onto a plane. \cref{fig:landmarks} shows all possible landmarks or field features: corners represented by \textit{L}, goal posts by \textit{G}, T-intersection by \textit{T}, and crosses by \textit{X}. Landmarks are described only by label and position relative to the robot (no orientation is inferred). Note that \cref{fig:landmarks} depicts the most ideal case where all the landmarks are observed without positional offsets, which will be referred to as noise throughout, and with no false-negative or false-positive features, which will be defined as false detections.

To analyze CLAP, a theoretical field is assumed to have the following properties:
\begin{itemize}
    \item Observed landmarks have no noise or false detections
    \item Landmarks have unique labels 
    \item The total number of landmarks is $l$, and
    \item All $l$ landmarks are observed.
\end{itemize}

\textit{Landmarks $l=0$}: 
In this case, no field features are observed. Thus, the robot cannot localize itself on the field with vision alone. Any state on the field is equally likely.


\textit{Landmark $l=1$}:
With a single field feature, the robot is bound to a circle with an infinite number of possible states encompassing that landmark.


\textit{Landmarks $l=2$}:
With distances to 2 field features, the robot's state can be limited to one of two possible positions. These positions are mirrored across the line connecting the two landmarks. Using the assumption that the landmarks are unique, however, only one of the two solutions remains valid. As can be seen by \cref{fig:mirroring}, each red player is constrained to a single side of the mirror (line connecting \textit{L} and \textit{T}) as the \textit{L} landmark must be to the robot's right and the \textit{T} landmark must be to the robot's left. This is a key point to make as relaxing the uniqueness of landmarks is vital to performing in the real world and will be handled later in this paper.



Given the two observed field features relative to the robot frame ($p_1^{\text{body}}$, $p_2^{\text{body}}$) and a set of matching features in world frame ($p_1^{\text{world}}$, $p_2^{\text{world}}$) from the a priori known map where the landmark orderings are the same (e.g.  \( p_1^{\text{world}} \) and \( p_1^{\text{body}} \) are both \textit{T}s), the robot's orientation \( \theta_{\text{robot}} \) can be calculated as:
\begin{equation*}
   \theta_{\text{robot}} = \text{atan2} (\Delta p_{y}^{\text{world}}, \, \Delta p_{x}^{\text{world}}) - \text{atan2} (\Delta p_{y}^{\text{body}}, \, \Delta p_{x}^{\text{body}}) 
\end{equation*}
\begin{equation}
    p_{robot} =\begin{bmatrix}
        x_{\text{robot}} \\ y_{\text{robot}}
    \end{bmatrix} = p_{1}^{\text{world}} - R(\theta_{\text{robot}})*p_1^{body}
\label{eq:mirror_match}
\end{equation}
Where $\Delta p = p_2 - p_1$, $p_{\text{robot}}$ is the robot's global position, and $R$ is a 2D rotation matrix.


\textit{$l > 2$ Landmarks}:
When more than two features are observed, each of the field features can be sampled as unique pairs, which results in $l\,\, choose\,\, 2$ combinations of
estimated states. In other words, the number of unique pairs formed with $l$ observations. Since each of these estimates only have one solution, each solution is coincident and lie exactly in the same position and orientation, resulting in a single state with the densest possible cluster. 

\textit{Non-Unique Landmarks}:
Upon relaxing the unique features assumption, each pair of landmarks may now have two possible states. If both landmarks are the same, the robot's relative perspective is indistinguishable when looking from the front or behind the mirror as can be seen by the blue players in \ref{fig:mirroring}. In this case, one of the estimated states is incorrect. If we denote $d$ as the number of pairs with the same field features, the total number of estimates can be calculated as such:

\begin{equation}
    E = \frac{l(l-1)}{2} + d.
    \label{eq:sum_N}
\end{equation}

Without $d$, the equation is simply the number of unique pairs. Assuming that all observed pairs match (discussed further in \cref{sec:matching}) to one of the correct field locations, $d$ also is the number of incorrectly estimated states. In practice, if there is a sufficient number of unique pairs observed, non-unique pairs can be ignored to reduce this effect.

Fortunately, since mirroring corresponds to a rotation by $\pi$, the orientation between the mirrored states will always be opposite of one another. Thus, each estimate will be unique. This avoids a singularity when trying to distinguish between correct and incorrect estimates in \cref{sec:clustering}.

\subsection{Landmark Matching}
\label{sec:matching}
\begin{figure}[t!]
    \centering
    \includegraphics[width=\linewidth]{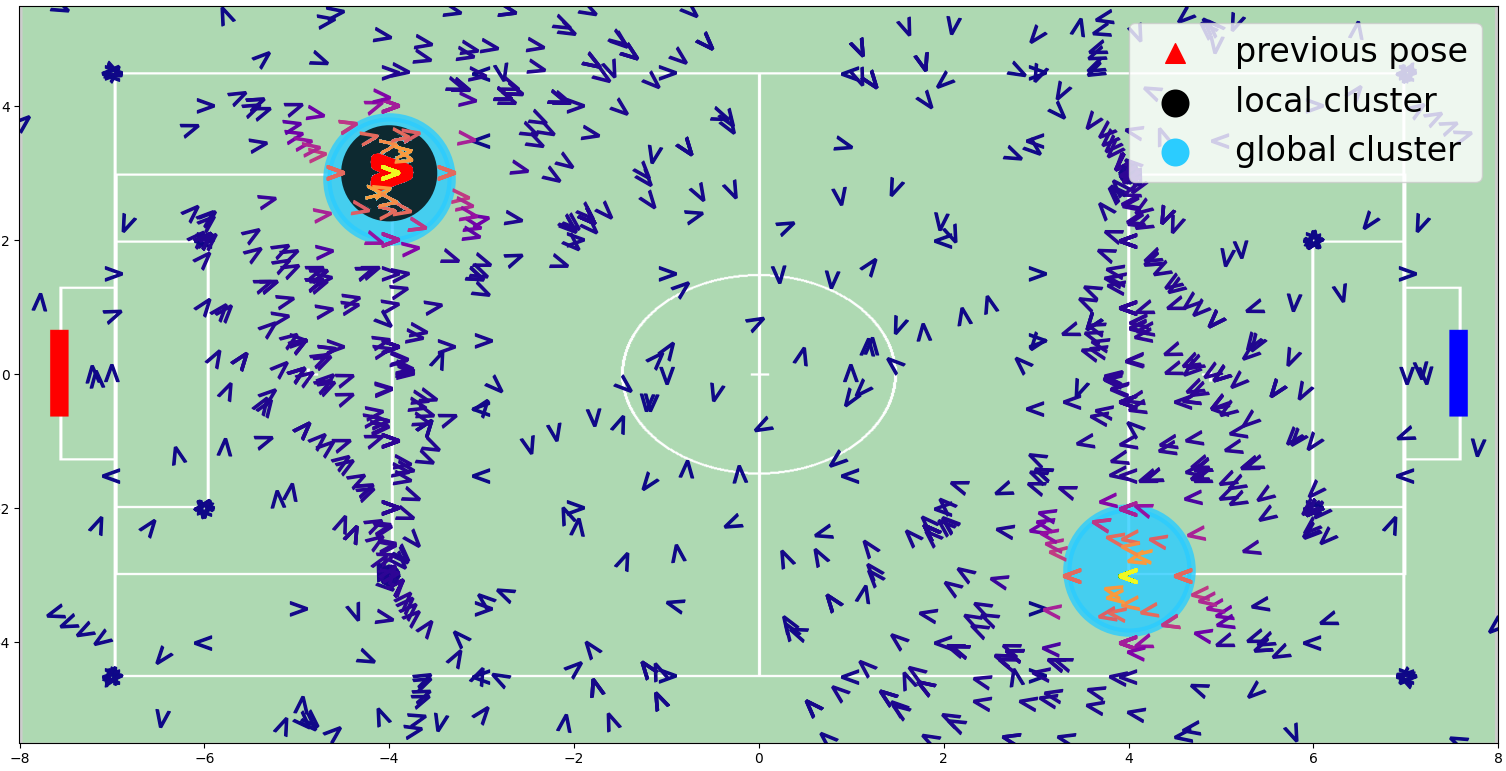}
    \caption{Depicts all estimated states, assuming we observe all the landmarks without noise and false detections. The orientation of the \textit{arrow} is the robot's orientation. The color of each estimation is colored to show density of similar points with \textit{dark blue} being low density and \textit{yellow} being high density. The number of points directly on top of each other at the yellow arrow is 595, and there are 2,278 points in total. The local and global clustering estimates are shown in \textit{black} and \textit{blue circles}. The previous pose, used as an initial guess for clustering, is depicted as a \textit{red arrow}.}
    \label{fig:cluster}
\end{figure}

On a full soccer field, multiple instances of a pair of landmarks with the same relative distance between them can exist. This makes them indistinguishable from each other when trying to match against the a priori map as shown by the red players in \cref{fig:mirroring}. Therefore, for every pair of landmarks used to generate an estimate, there can exist as many incorrect estimates as there are instances of that pairing. For example, if there are 4 total similar pairs, then there would be 4 estimated states with 1 of them being correct and the others incorrect. Consequently, the number of incorrect estimates increases with less number of unique features and an increase in noise, while the number of correct estimates always remains constant. However, since geometric mirroring is being matched to a global map with relatively distanced features, the generated estimated states spread out across the field in terms of position as well as orientation in many cases due to the $\pi$ rotation from mirroring resulting in the left and right set of states being similar as shown in \cref{fig:cluster}. This will be resolved in \cref{sec:clustering}.

\textit{Noise}:
As the noise in the relative position of the landmarks increases, matching becomes far more difficult due to field features becoming indistinguishable between other similar pairs. It is unknown if the distance between landmarks is due to the true distance or presence of noise. Since each distance between the pairs of landmarks constitutes an identifier for the corresponding pairs in the a priori known global map of the field, noisy distance measurements begin to look like other identifiers. 

For example, if our robot observes a \textit{T} and an \textit{L} landmark, then it could be any pair of \textit{T}s and \textit{L}s in the corresponding global map as shown in \cref{fig:landmarks}. By a priori knowing the distances between each landmark, one could reduce the number of possibilities drastically if the noise is small enough. 
The dimension of the field within the white lines is 14 m by 9 m. If the distance between the observed \textit{T} and \textit{L} is 7 m without noise, then the \textit{T} and \textit{L} have to be 1 of the 4 \textit{T} and \textit{L} pairs that span half the field along the length marked with white lines. If the magnitude of noise reached to be $\pm 20$ m, all \textit{T} and \textit{L} combinations would be possibilities since it would be impossible to determine whether the distance was due to noise or not. Instead of a single value as an identifier, a range of values are used. The lower the noise, the smaller the range will be. The higher the noise, the larger the range will be, subsuming all identifying distances.



The process of estimating states based on landmarks remains the same regardless of whether or not there is noise. The only difference is the number of potentially similar pairs of features in the a priori map could increase the number of faulty matches, increasing the number of incorrect estimates.





\textit{Noiseless \& False Detections}:
False negatives are treated as missed observations and have no bearing on incorrect state estimates. False positives are more consequential since they are inconsistent with the a priori global map, meaning that the estimate will be incorrect since the relative distance will be matched to an incorrect feature. However, there is a high chance that the incorrect estimated states will spread out as seen in \cref{fig:cluster}, since the relative distances to the robot and the features need to be consistent over several camera observations for multiple false positives to cluster together. 



The number of incorrect states due to false positives can be derived similarly to that of the unique pairs term in \cref{eq:sum_N}, assuming unique landmarks. Given there are $m$ false positive detections then number of incorrect pairs, $M$, are:
\begin{equation}
\centering
    \begin{split}
        M &= (l-1)+(l-2)+\dots+(l-m)
        =ml-\sum\limits_{\alpha=1}^m\alpha \\
        &=\frac{2ml-m^2-m}{2}
    \end{split}
\end{equation}
As $m \rightarrow l$, $M$ approaches the first term in \cref{eq:sum_N}. The number of incorrect estimates could scale similarly to that of good estimates. Fortunately, false positives do not happen often (see \cref{sec:results}) and the incorrect estimates spread out.

\subsection{Clustering}
\label{sec:clustering}
\cref{sec:geo_mirror} and \cref{sec:matching} detailed several cases where incorrect state estimates occurred due to the symmetry of the field, faulty landmark matches, and faulty detection. Two solutions from the same field features can be controlled by not using those pairs. False detections do not occur frequently if the vision model is properly trained. However, matching alone results in a larger number of incorrect estimates than correct ones, which poses a problem.




Fortunately, incorrect estimates tend to spread out relative to its position and orientation as seen in \cref{fig:cluster}. Geometric mirroring across the same field features results in a $\pi$ rotation across its orientation, meaning that even though its position experiences a singular point, the full state does not. Additionally, the field features are spread over large distances. This means that for matching the incorrect matches will be equally spread. Finally, even with noise for similar states, to densely cluster together requires consistency with the a priori known global feature map.

The spread of estimates is seen in \cref{fig:cluster}. In this instance there are 2,278 potential states. The color map shows the density of similar points in both position and orientation. The more similar estimates are within a given estimates area the brighter the color with \textit{yellow} being the brightest. The two brightest \textit{yellow} arrows, surrounded by bright blue circles, had 595 estimates directly stacked on top of each other. That means roughly 52\% of all points were in these two spots. 



 As seen in \cref{fig:cluster}, correct state estimates are consistent and tend to group together. This leads to the natural conclusion of using a clustering method. 
There are many different clustering algorithms: approaches based on splitting and merging like ISODATA \cite{ball1964some,jain1988algorithms}, density-based clustering method like DBSCAN \cite{khan2014dbscan}, model-based clustering like GMM \cite{dempster1977maximum} and mean shift-based clustering \cite{fukunaga1975estimation,cheng1995mean}
In our case we used $K$-means clustering \cite{macqueen1967some,forgy1965cluster,ismkhan2018ik}. We chose $K$-means because it is fast, and with enough observed field features we can consistently identify 2 centroids due to the two-fold symmetry of soccer fields.







\textit{Na\"ive Local Clustering}:
To reduce computational load and increase solving speed a local na\"ive clustering approach was used by only considering a single cluster within a certain distance to its last estimated state, as depicted by the black ring in the upper left penalty box in \cref{fig:cluster}. In practice, this works surprisingly well, where in many cases even with high levels of noise it would correct itself, meaning there was a relatively large \textit{bowl of attraction} for the state to be perturbed. The \textit{bowl of attraction} can also be seen in \cref{fig:cluster} by the area of lighter colored arrows formed around the deepest part of the bowl, the yellow arrow. In this method, estimates are averaged in the cluster to get the next estimate. This means that if the cluster was not at the center of the bowl surrounding estimates would pull it in that direction.



\textit{Global Clustering}:
For redundancy and safety purposes, we have a slower clustering method taking into account all estimated states over several frames. Estimates from older frames are propagated using IMU estimates since drift is negligible for short intervals. This global clustering method checks the na\"ive approach at a slower rate, and if they do not agree within some threshold the pose gets reset to where this method estimates it to be. This state \textit{reset} is represented by the \textit{Merge Estimates} box in \cref{fig:framework}.

Given the symmetry of a soccer field, the global clustering can limit the state to be in one of two positions, meaning that it can correct up to at most half the field. We use the previous state to pick the centroid closest to that state.

An additional problem arises with the vast majority of states being outliers, centroids from $K$-means can be heavily influenced. To alleviate this issue, we iteratively run the $K$-means algorithm \cite{harale2011iterative}, while removing some percentage of outliers which are far from the centers in each iteration to refine its estimate. \cref{alg:pseudocode} gives an overview of how CLAP works and where clustering plays a role.







\begin{algorithm}[t!]
\caption{CLAP}
\label{alg:pseudocode}
\begin{algorithmic}[1]
\State \textbf{Input:} Observations \(\{p_1^{\text{body}} \cdots \, p_n^{\text{body}}\}\) and Estimate $\mathbf{r}_{t-1}$
\State \textbf{Output:} \( \mathbf{r}_{t} = (x_{\text{robot}, t}, y_{\text{robot}, t}, \theta_{\text{robot}, t})\)
\State \textbf{Initialize:} Lists $\mathbf{A}$, \(\mathbf{C}\) and Thresholds $\delta_t$, $\delta_c$
\For{Each Combination \((p_i^{\text{body}}, \, p_j^{\text{body}} )\)} 
    \For {Each Similar Apriori Map Pair \((p_i^{\text{world}}, \, p_j^{\text{world}})\)} 
        \State \(\mathbf{r}_{t, k}\coloneqq f((p_i^{\text{body}}, \, p_j^{\text{body}} ), (p_i^{\text{world}}, \, p_j^{\text{world}}))\), \cref{eq:mirror_match}
        \State Append $\mathbf{r}_{t, k}$ to \(\mathbf{A}\) 
        \If { \(\|\mathbf{r}_{t, k} - \mathbf{r}_{t-1} \|< \delta_t\)}
            \State Append $\mathbf{r}_{t, k}$ to \(\mathbf{C}\)
        \EndIf
    \EndFor
\EndFor
\State \(\mathbf{r}_{\text{t, L}} \coloneqq \frac{\sum C}{|C|}\), see \cref{sec:clustering}
\State $\mathbf{r}_{\text{t, G}} \coloneqq K$-means$(\mathbf{A})$ Centroid Nearest to $\mathbf{r}_{t-1}$
\If{ $\| \mathbf{r}_{\text{t, L}} - \mathbf{r}_{\text{t, G}} \| > \delta_c$}
    \State $\mathbf{r}_{t} \coloneqq \mathbf{r}_{\text{t, G}}$
\Else
    \State $\mathbf{r}_{t} \coloneqq \mathbf{r}_{\text{t, L}}$
\EndIf
\end{algorithmic}
\end{algorithm}

\subsection{Other Components} 
\label{sec:other_components}

\textit{IMU Estimation}: 
The robot state is also estimated using IMU data and foot contact sensor measurements, using a similar approach to that of the Contact-Aided Invariant Extended Kalman Filter (InEKF) \cite{hartley2019contactaidedinvariantextendedkalman}. However, because of the high impact forces on each touchdown and the absence of magnetometer aiding, the yaw angle relative to the field frame significantly drifts over time. 

\textit{Particle Filter}: 
CLAP can work by itself; however, there are times when the robot is looking down to kick the ball and does not see any landmarks. This results in inconsistent jumpy estimates when using CLAP as the sole localizer.
Therefore, we also use a particle filter to fuse data from the InEKF to produce a smoother estimate for higher-level planning, as shown in \cref{fig:framework} \cite{chen2010particle,tariq20242d}. In the prediction step, the particle filter uses the InEKF's position and rotation matrices to calculate velocity, which is used to propagate the particles. When receiving a state estimate from CLAP, it updates the weight of each particle. In every loop, the particle filter estimates the state by taking a weighted average of the particles. Since CLAP updates less frequently than the IMU due to the camera rate, the robot only localizes with the IMU in between CLAP estimates.

\section{Results} 
\label{sec:results}

\subsection{Experimental Setup}



During the competition, the ZED 2i was used to capture images and estimate depth of pixels from stereo vision, in compliance with the RoboCup rules. For vision processing, YOLO-v8 was used to detect field features \cite{yolov8,dwijayanto2019real,kim2023enhancing}. The output rate of the ZED 2i and YOLO-v8 model averaged about 40 Hz. Meanwhile, the Microstrain 3DM-CV7-AHRS was used for the IMU with the InEKF estimating position, orientation, velocity, and acceleration at about 500 Hz. All code ran on an HP Elite Mini 800 G9, which contained an Intel Core i7-12700T and an Nvidia GeForce RTX 3050 Ti 4GB \cite{ahn2023}. Both the localization and particle filter ROS 2 (robot operating system) nodes were capped at 100 Hz and were able to consistently stay at that rate, even when observing multiple landmarks or running the $K$-means clustering. In order to reduce lag spikes, the max number of landmarks from the YOLO-v8 node was limited to 7. The vision module had a distance-to-percentage error on the position of landmarks of: $<2m$ to $\pm 1 \%$, $<4m$ to $\pm 3 \%$, $<6 m$ to $\pm 5 \%$, $<8m$ to $\pm 8\%$, and $>8m$ to $\pm 10 \%$, which is fairly linear. The false detection rate on average was around $3\%$ of total field features. Additionally, the robot’s ground truth pose is measured using a Vicon motion capture (mocap) system operating at $300 Hz$.

\subsection{Validation}
\textit{Simulation} We simulated a rectangular trajectory where the vertices are the corners of the goal box. Uniform noise of \(\pm 0.5 m\) was added to landmark observations, and uniform noise of \(\pm 0.02 m\) for position and \(\pm0.02 rad\) for orientation was added to the IMU estimation every \(10 ms\). The field of view in this simulated test was \(110\degree \) which is the same as the ZED 2i camera used in competition.

\begin{figure}[htbp]
    \centering
    \begin{subfigure}[b]{0.23\textwidth}
    \includegraphics[width=\textwidth]{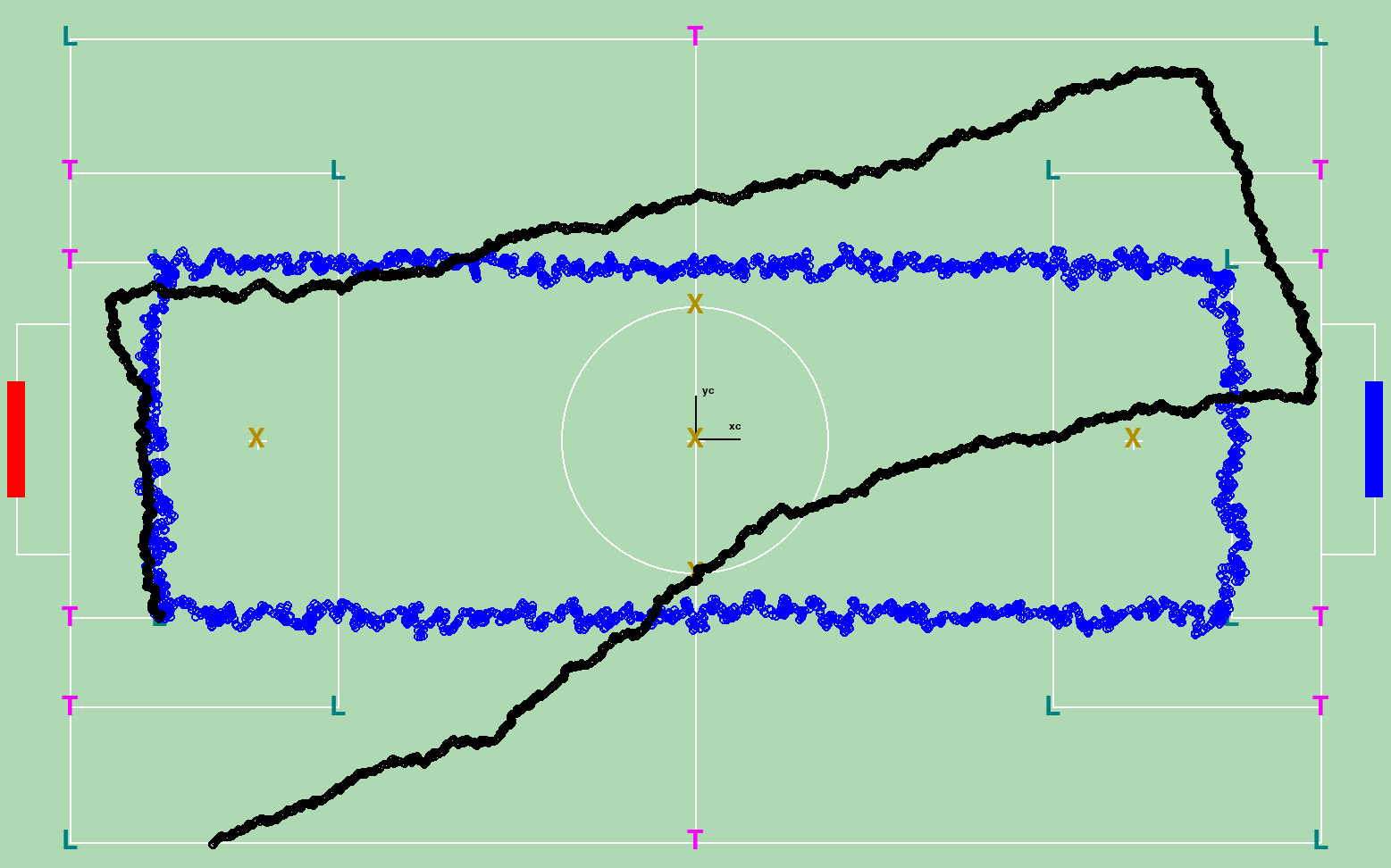}
    \caption{}
    \label{fig:CLAP_VS_IMU}
    \end{subfigure}
    \hfill
    \begin{subfigure}[b]{0.23\textwidth}
        \includegraphics[width=\textwidth]{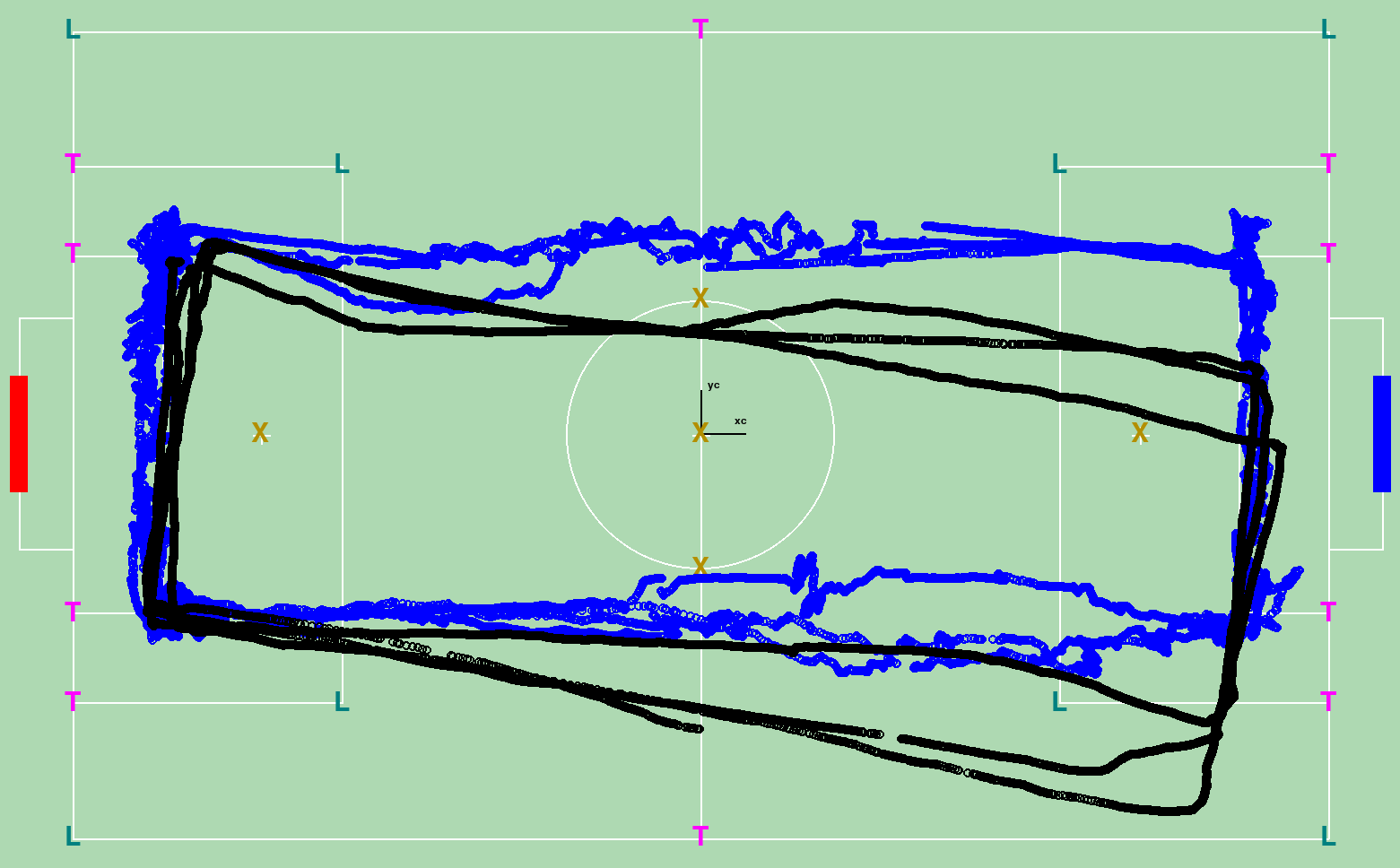}
    \caption{}
    \label{fig:CLAP_VS_IMU_real}
    \end{subfigure}
    \caption{(a) Simulated box trajectory comparing CLAP and IMU-only and (b) real trajectory. The CLAP output is shown in \textit{blue} and the IMU-only output is in \textit{black}.}
\end{figure}

\cref{fig:CLAP_VS_IMU} shows our estimated trajectory by our localization method with CLAP in blue and an estimated trajectory only using IMU data in black. Since this was simulated, other components did not computationally bottleneck localization, allowing it to run at \(+250 Hz\). 
The position error was on the order of \(0.1 m\) for global positioning and \(0.02 rad\) for orientation. Given the magnitude of noise these results seem reasonable.

\textit{Real World}: In \cref{fig:CLAP_VS_IMU_real}, we manually controlled our robot to walk along a trajectory as close to \cref{fig:CLAP_VS_IMU} as possible. We ran it for a total of three laps at a speed around \(0.3 m/s\). Its trajectory stayed relatively close to the trajectory in \cref{fig:CLAP_VS_IMU} despite real-world disturbances. The estimate with just the IMU alone continues to drift over time. Although ground truth positions could not be obtained due to the limited coverage area of the motion capture system, preventing error statistics as in simulation, this experiment demonstrates CLAP’s capability for long-distance localization across the entire soccer field.

\begin{figure}[t!]
    \centering
    \includegraphics[width=\linewidth]{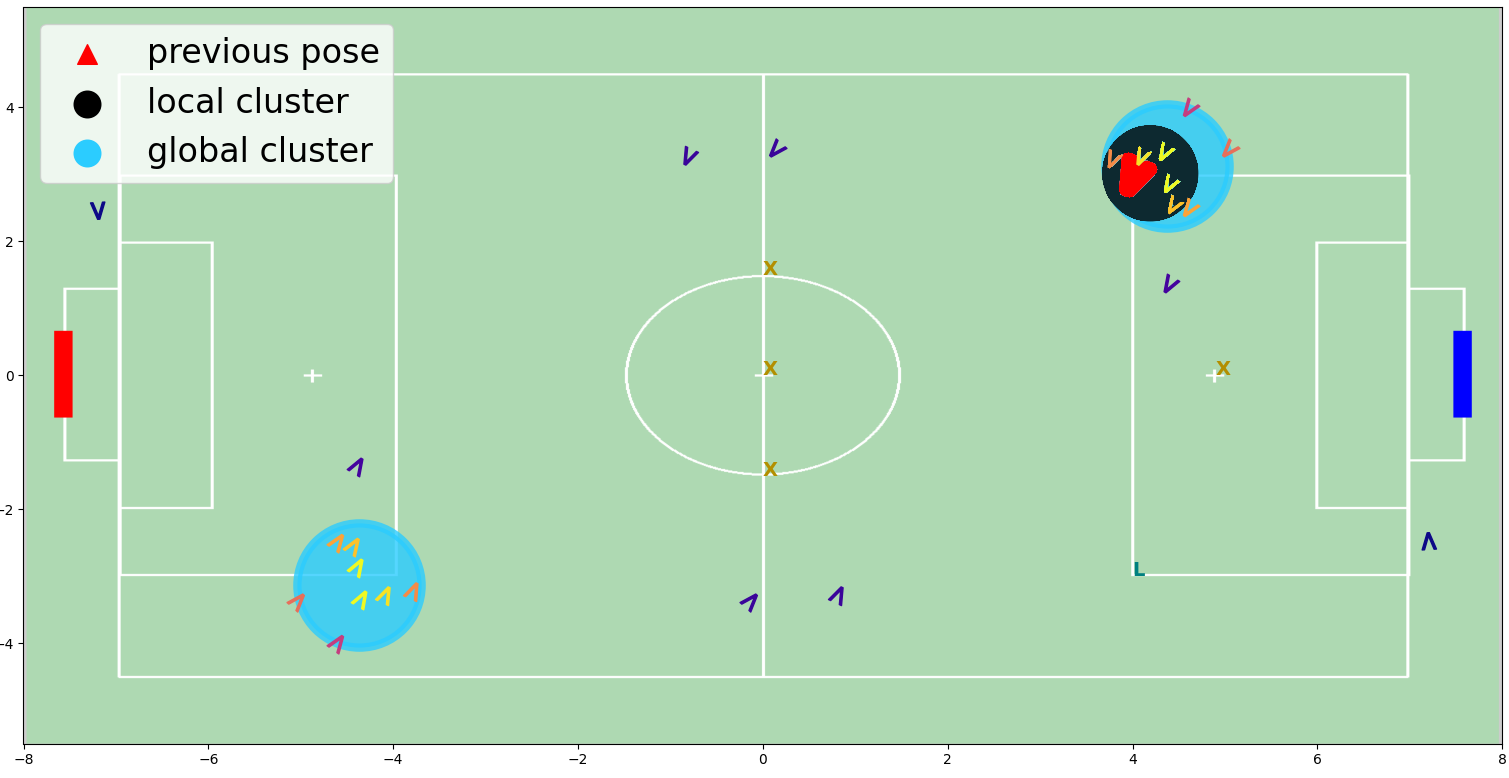}
    \caption{Real localized state at the center of the \textit{black circle} on our robot with estimated states similar to \cref{fig:cluster}. Notice that in this instance only a subset of landmarks were observed and experience noise such that their positions in the figure are incorrect. In total there are 26 estimates with 9 estimates in each cluster.}
    \label{fig:real_cluster}
\end{figure}

\cref{fig:real_cluster} shows what our robot, depicted as a red triangle,  may see in a single camera frame. Note our robot only sees a subset of landmarks and noise moves them from their exact positions. This results in only 26 estimates in contrast to \cref{fig:cluster}. Despite fewer estimates, two prominent clusters form with surrounding estimates that maintain a very similar orientation, ie, the \textit{ bowl of attraction}. The local cluster, shown as a black circle is contained within the global cluster in bright blue. The global cluster considers more points so it is liable to be slightly shifted towards incorrect states, however, no resetting occurs because the global and local cluster are still close to each other.

\begin{figure}[htbp]
    \centering
    \begin{subfigure}[b]{0.15\textwidth}
        \includegraphics[width=\textwidth]{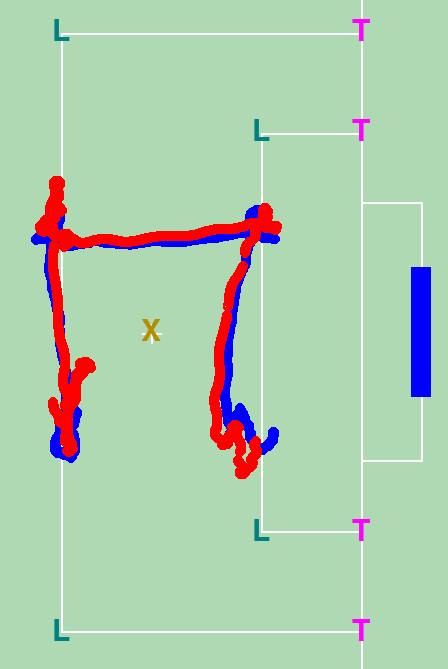}
        \caption{}
        \label{fig:a}
    \end{subfigure}
    \hfill
    \begin{subfigure}[b]{0.15\textwidth}
        \includegraphics[width=\textwidth]{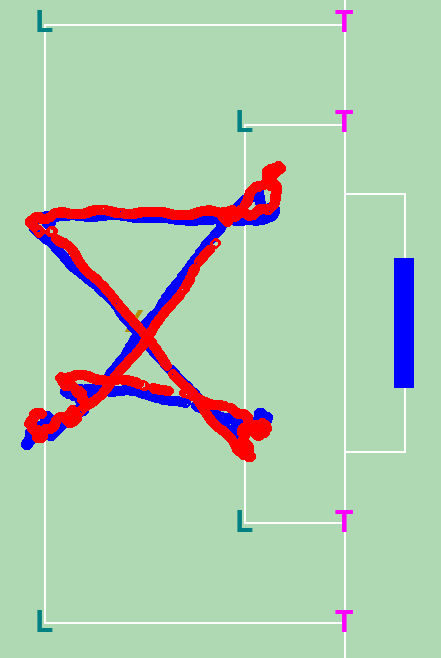}
        \caption{}
        \label{fig:b}
    \end{subfigure}
    \hfill
    \begin{subfigure}[b]{0.15\textwidth}
        \includegraphics[width=\textwidth]{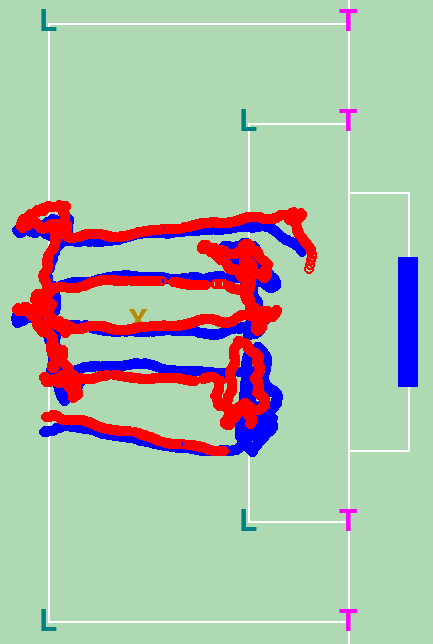}
        \caption{}
        \label{fig:c}
    \end{subfigure}
    \caption{Three trajectories recorded on a real soccer field. Only a single lap of the whole trajectory is pictured for clarity. The localization results from CLAP are shown in \textit{blue}, while the ground truth positions obtained from the Vicon motion capture system are shown in \textit{red}: (a) C-shaped, (b) X-shaped, and (c) Zigzag. }
    \label{fig:threeimages}
\end{figure}

\textit{Accuracy}:
\cref{fig:threeimages} show three localization trajectories with CLAP in \textit{blue} and ground truth using a Vicon motion capture system in \textit{red}. For clarity, only a single lap of the whole trajectory is shown. Due to the limitations of our motion capture setup, we could only test in a square region in the right half of the soccer field shown by the trajectories in \cref{fig:threeimages}. In certain positions, the motion capture system has trouble seeing the markers on the robot resulting in jumps in the mocap data. These jumps will not be considered in future accuracy calculations.

\cref{tab:error_statistics_v2} shows the mean absolute error (MAE) and the standard deviation (std) between the localization output and the ground truth. In addition to CLAP, we also ran the same 3 trajectories using ILM and aMCL. All three of these methods consist of an IMU-estimation component and a land-mark-matching component. ILM has an additional outlier-dropping component that uses the RANSAC algorithm. We decided to run ILM without the RANSAC component to focus on comparing the matching method of ILM, which is its modified form of ICP, with CLAP and aMCL. RANSAC in ILM is just a robust estimation tool and could theoretically be added to all these  as a post-processing measure. All three methods ran at around $100 Hz$ for comparison. As shown in \cref{tab:error_statistics_v2}, the error in CLAP is comparable to other methods.

\begin{table}[htbp]
    \centering
    \begin{subtable}[t]{\linewidth}
        \centering
        \begin{tabular}{l|ccc}  
            \toprule
            \multirow{2}{*}{Trajectory} & \multicolumn{3}{c}{\makecell{Position MAE $\pm$ std (meter)}} \\
            \cmidrule(lr){2-4} 
            & CLAP & ILM w/o RANSAC & aMCL  \\
            \midrule
            Trajectory 1 & 0.215$\pm$0.136 & 0.173$\pm$0.128 & 0.175$\pm$0.166    \\
            Trajectory 2 & 0.195$\pm$0.109 & 0.168$\pm$0.091 & 0.242$\pm$0.167      \\
            Trajectory 3 & 0.178$\pm$0.093 & 0.165$\pm$0.085 & 0.219$\pm$0.125      \\
            \bottomrule
        \end{tabular}
        \caption{Position Error Statistics}
        \label{tab:position_error}
    \end{subtable}

    \vspace{1em}

    \begin{subtable}[t]{\linewidth}
        \centering
        \begin{tabular}{l|ccc}  
            \toprule
            \multirow{2}{*}{Trajectory} & \multicolumn{3}{c}{\makecell{Orientation MAE $\pm$ std (degree)}} \\
            \cmidrule(lr){2-4}
            & CLAP & ILM w/o RANSAC & aMCL \\
            \midrule
            Trajectory 1 & 3.138$\pm$1.631 & 2.461$\pm$2.029 & 2.329$\pm$2.312     \\
            Trajectory 2 & 2.925$\pm$2.423 & 3.199$\pm$2.552 & 3.247$\pm$2.660    \\
            Trajectory 3 & 2.875$\pm$2.317 & 2.411$\pm$2.152 & 2.312$\pm$1.931    \\
            \bottomrule
        \end{tabular}
        \caption{Orientation Error Statistics}
        \label{tab:orientation_error}
    \end{subtable}
    \caption{Mean Absolute Error (MAE) and one standard deviation for different trajectories comparing CLAP, ILM without RANSAC, and aMCL.}
    \label{tab:error_statistics_v2}
\end{table}

\begin{figure*}[htbp!]
    \centering
    \includegraphics[width=0.9\linewidth]{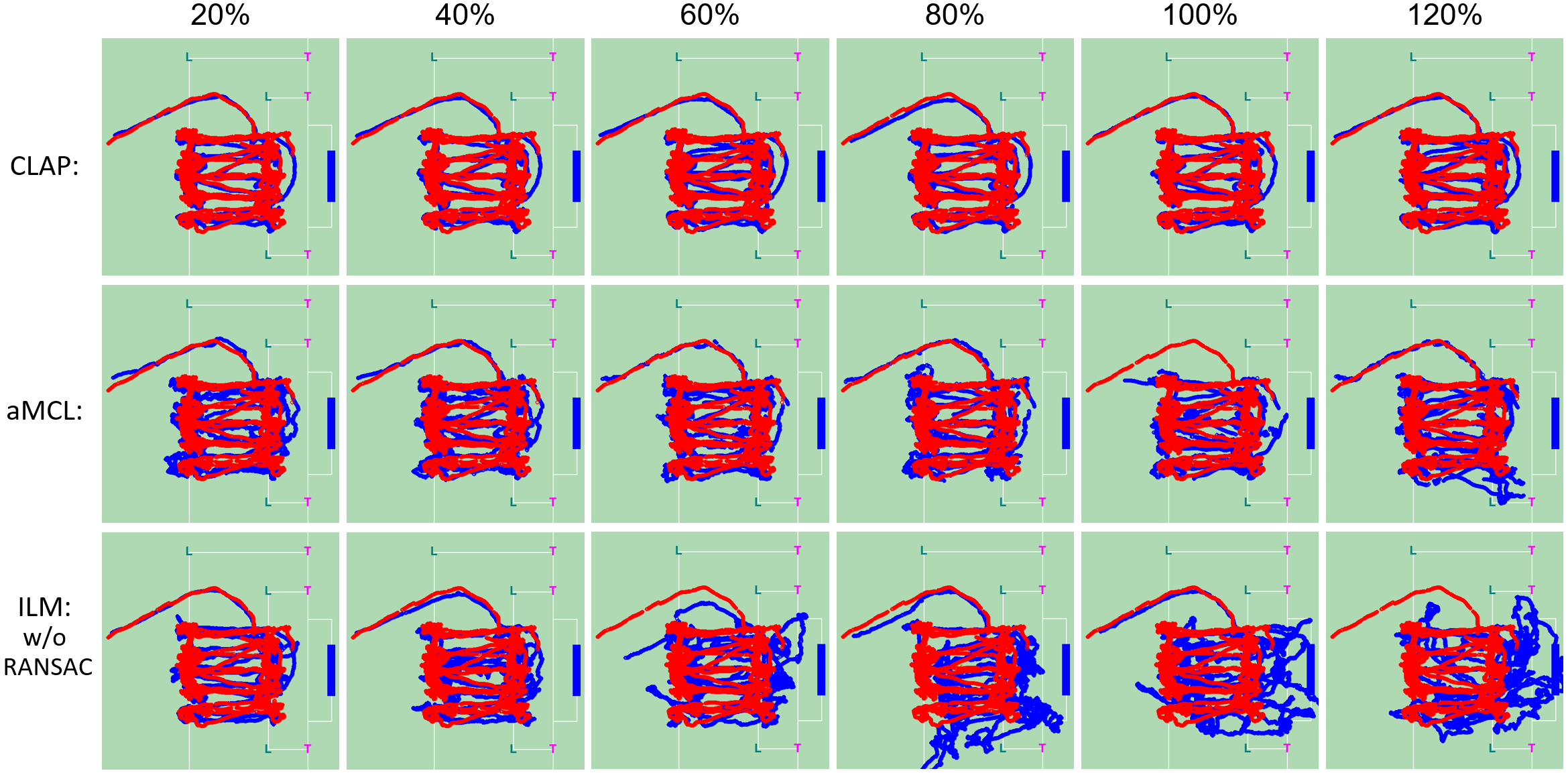}
    \caption{Comparing the localization results of trajectory 3 in full from CLAP, aMCL, and ILM without RANSAC when subjected to false landmarks. As opposed to \cref{fig:c}, the entire trajectory is shown. The motion capture data is shown in \textit{red} and the localization output is shown in \textit{blue}. The percentages indicate the ratio of false landmarks to real landmarks. CLAP remains consistent and close to the motion capture throughout all trials.}
    \label{fig:compare_methods_outliers_traj3}
\end{figure*}

\textit{Robustness}:
Robustness tests were conducted where random false landmarks (outliers) were appended to the robot's observations (inliers) while walking in similar trajectories to \cref{fig:threeimages} but with more laps. Trajectory 1, 2, and 3 took $272s$, $326s$, and $363s$, respectively. A single lap is shown in \cref{fig:threeimages} for clarity. The ratio of false landmarks to real landmarks ranged from 20-120\%. This means that for a 100\% ratio, if a robot sees 5 correct landmarks, 5 incorrectly-placed landmarks will be appended to the robot's observations. While 120\% may be extreme, we conducted this experiment because we observed our opponents' robots localization suddenly flipping and kicking the ball into their own goal at competition. If the number of observed features is small a false detection would make up a large percentage. Additionally, faded field lines and neighboring fields would all lead to numerous false detections.
\cref{fig:compare_methods_outliers_traj3} shows the results of the entire trajectory 3 where CLAP maintained a consistent output between trials and a close output to the motion capture data in all the tests while ILM and aMCL began diverging as the number of false detections grew. The other two trajectories will be included in the supplementary video, but have similar results.

\begin{figure}[htbp]
    \centering
    \begin{subfigure}[b]{0.15\textwidth}
        \includegraphics[width=\textwidth]{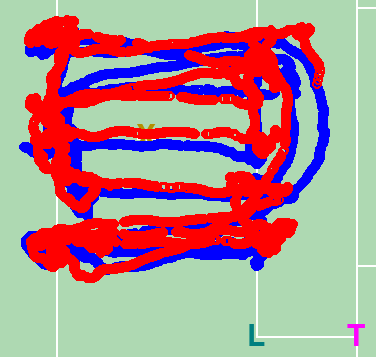}
        \caption{}
        \label{fig:a_r}
    \end{subfigure}
    \hfill
    \begin{subfigure}[b]{0.15\textwidth}
        \includegraphics[width=\textwidth]{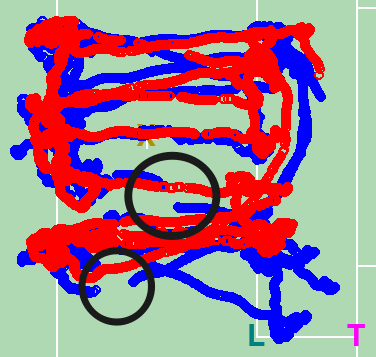}
        \caption{}
        \label{fig:c_r}
    \end{subfigure}
    \hfill
    \begin{subfigure}[b]{0.15\textwidth}
        \includegraphics[width=\textwidth]{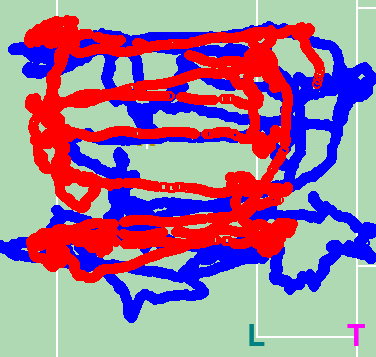}
        \caption{}
        \label{fig:b_r}
    \end{subfigure}
    \caption{Closeup results of \cref{fig:compare_methods_outliers_traj3} with 120\% more false landmarks than real landmarks appended to the robot's observations on three different localization methods (a) CLAP, (b) aMCL, and (c) ILM without RANSAC. Only a portion of the mocap trajectory is used to emphasize the gaps. The localization result is \textit{blue}, while the ground truth position from motion capture is in \textit{red}. The \textit{black circles} show where the aMCL method jumps.}
    \label{fig:three_traj_robust_120}
\end{figure}

\cref{fig:three_traj_robust_120} shows a closeup of the 120\% false detections trial in \cref{fig:compare_methods_outliers_traj3}. To clearly show the jumps, a subsection of the full trajectory was used. The poses from CLAP are consistent while aMCL has noticeable gaps between poses, as circled in black, which show that the localization is jumping around.

\begin{figure}[htbp]
    \centering
    \includegraphics[width=0.85\linewidth]{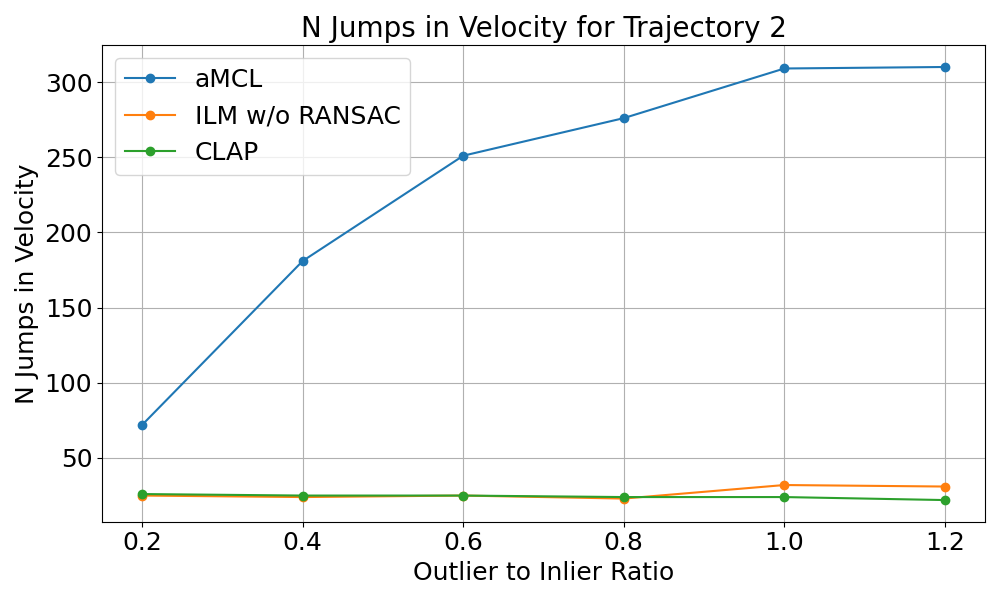}
    \caption{Number of velocity jumps in a similar but longer path to trajectory 2 of \cref{fig:threeimages}. A velocity jump is characterized by a velocity greater than $16\,\mathrm{m/s}$ in linear speed and $4\,\mathrm{rad/s}$ in angular speed.}
    \label{fig:vel_jump}
\end{figure}

\begin{figure}[htbp]
    \centering
    \includegraphics[width=0.85\linewidth]{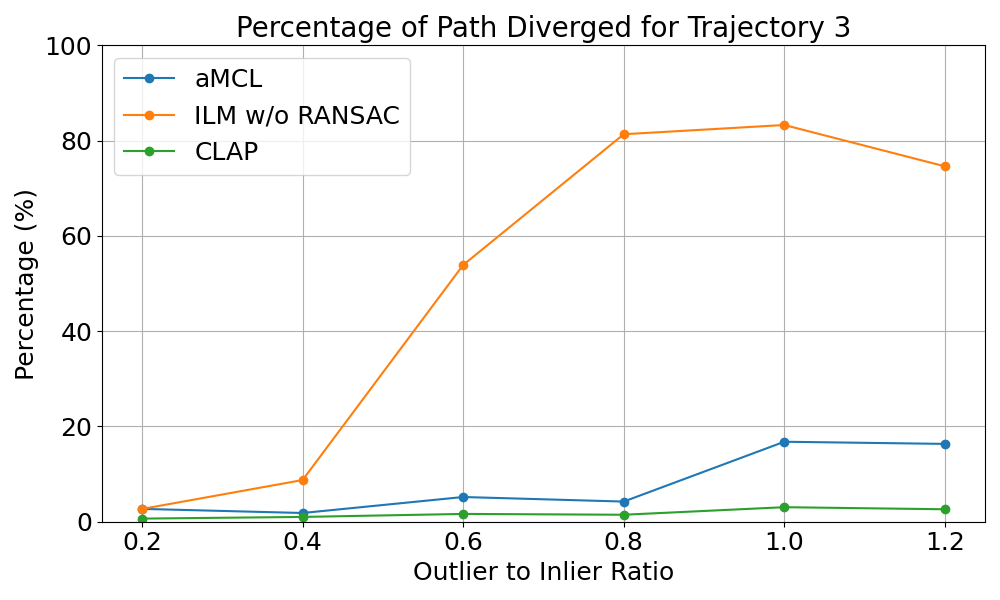}
    \caption{Percentage of the total path that has diverged from the motion capture ground truth. The trajectory is the same as in \cref{fig:compare_methods_outliers_traj3}. Divergence is characterized by a difference greater than $0.5 m$ in position and $0.15rad$ in orientation from the motion capture ground truth.}
    \label{fig:divergence}
\end{figure}

We can quantify how "jittery" each localization is by searching for abrupt changes in the robot's pose. We define a velocity jump as a velocity greater than $16\,\mathrm{m/s}$ and $4\,\mathrm{rad/s}$ in both linear and angular speed, respectively. 
\cref{fig:vel_jump} shows the results of the robot proceeding in a trajectory similar to \cref{fig:b}, but with more laps. The graph shows that in all trials, CLAP had 22-25 jumps, ILM had 23-32 jumps, and aMCL had 72-310 jumps. At high outlier-to-inlier ratios, aMCL jumps over $1\%$ of the time which corresponds to roughly 1 jump per second, whereas CLAP and ILM jump once about every $12s$.  However, as \cref{fig:compare_methods_outliers_traj3} indicates, ILM is more susceptible to divergence. \Cref{fig:divergence} shows the percentage of the trajectory in \cref{fig:compare_methods_outliers_traj3} that deviates more than $0.5\,\mathrm{m}$ in position or $0.15\,\mathrm{rad}$ in orientation from the motion capture ground truth. For outlier-to-inlier ratios greater than 40\%, ILM spends over half of its trajectory diverging.

In some preliminary tests of ILM with RANSAC against CLAP using Python, we found that RANSAC reduced ILM to \(15 Hz\) and diverged similarly to \cref{fig:compare_methods_outliers_traj3}. Rewriting ILM and RANSAC in C++ showed robustness and accuracy similar to CLAP in Python, but further testing needs to be done.
\section{Conclusion} 
\label{sec:conclusion}
In this paper, we introduced a novel method for estimating states from pairs of feature observations and then clustering them to localize, called CLAP. Theoretical intuition of CLAP was presented, followed by simulation and real-world field experiments. Then it was deployed on the field for the Robocup competition, where localization played a crucial role in scoring the 45 goals in the seeded matches and the 6 in the championship game \cite{symposium}. Many of our goals were taken from up to 4 meters away from the goal and off-center, requiring accurate and steady-state estimation to aim. In the championship match, one of our robots even maneuvered behind the other within the goal, which is only 0.6 m deep, to clear a ball a few centimeters from going into our goal. It also proved to be more robust than competitors whose localization flipped and scored in their own goal or others who got lost due to field features from adjacent fields. 

\textit{Future Work}: CLAP can be further improved by parallelizing its geometric calculations to run independently. Separately, for more general purposes, this can be expanded to 3 dimensions by simply projecting to a horizontal plane. Further comparisons between CLAP and ILM with RANSAC on speed and robustness will also have to be made. In addition, the effects of false positives in asymmetric cases will have to be tested, as the false positives may introduce false symmetries where more than one possibility exits. CLAP can also be applied to other problems involving humanoid robots navigating in 2D settings with known maps, such as warehouse or factory environments.

\end{document}